\documentclass[conference]{IEEEtran}
\IEEEoverridecommandlockouts
\usepackage{cite}
\usepackage{amsmath,amssymb,amsfonts}
\usepackage{algorithmic}
\usepackage{graphicx}
\usepackage{textcomp}
\usepackage{xcolor}

\usepackage{enumitem}
\usepackage{booktabs}
\usepackage{svg}
\usepackage{cleveref}

\usepackage{titlesec}
\def\BibTeX{{\rm B\kern-.05em{\sc i\kern-.025em b}\kern-.08em
 T\kern-.1667em\lower.7ex\hbox{E}\kern-.125emX}}

\usepackage{fancyhdr,lipsum}
\fancypagestyle{firstpage}{
  \fancyhf{}
  \fancyhead[C]{To appear at the 2023 International Joint Conference on Neural Networks (IJCNN), Queensland, Australia, June 2023.}
  \fancyfoot[C]{\thepage}
}

\begin{document}

\title{RobCaps: Evaluating the Robustness of Capsule Networks against Affine Transformations and Adversarial Attacks}


\author{\IEEEauthorblockN{Alberto Marchisio\textsuperscript{1,*}\thanks{*These authors contributed equally to this work.}, Antonio De Marco\textsuperscript{2,*}, Alessio Colucci\textsuperscript{1}, Maurizio Martina\textsuperscript{2}, Muhammad Shafique\textsuperscript{3}}
\IEEEauthorblockA{\textit{\textsuperscript{1}Technische Universit{\"a}t Wien, Vienna, Austria}\ \ \ \textit{\textsuperscript{2}Politecnico di Torino, Turin, Italy}\ \ \ \textit{\textsuperscript{3}New York University, Abu Dhabi, UAE}} 
\IEEEauthorblockA{\textit{Email: alberto.marchisio@tuwien.ac.at, s254593@studenti.polito.it, alessio.colucci@tuwien.ac.at}}
\IEEEauthorblockA{\textit{maurizio.martina@polito.it, muhammad.shafique@nyu.edu}}\\
\vspace*{-40pt}}

\maketitle
\thispagestyle{firstpage}

\begin{abstract}
Capsule Networks (CapsNets) are able to hierarchically preserve the pose relationships between multiple objects for image classification tasks. Other than achieving high accuracy, another relevant factor in deploying CapsNets in safety-critical applications is the robustness against input transformations and malicious adversarial attacks.

In this paper, we systematically analyze and evaluate different factors affecting the robustness of CapsNets, compared to traditional Convolutional Neural Networks (CNNs). Towards a comprehensive comparison, we test two CapsNet models and two CNN models on the MNIST, GTSRB, and CIFAR10 datasets, as well as on the affine-transformed versions of such datasets. With a thorough analysis, we show which properties of these architectures better contribute to increasing the robustness and their limitations. Overall, CapsNets achieve better robustness against adversarial examples and affine transformations, compared to a traditional CNN with a similar number of parameters. Similar conclusions have been derived for deeper versions of CapsNets and CNNs. Moreover, our results unleash a key finding that the dynamic routing does not contribute much to improving the CapsNets' robustness. Indeed, the main generalization contribution is due to the hierarchical feature learning through capsules.
\end{abstract}


\section{Introduction}

In recent years, many works have explored the problems of adversarial examples and affine transformations in Convolutional Neural Networks (CNNs) for image classification applications~\cite{Capra_2020Access_HWSWDNNSurvey}\cite{Marchisio_2019ISVLSI_DL4EC}\cite{Capra_2020MDPI_UpdatedSurvey}\cite{Dave_2022VTS_SS}. 
Szegedy et al.~\cite{szegedy} proposed the concept of adversarial examples, i.e., examples with small perturbations, imperceptible to the human eye, that mislead high-confidence models when added to the input. 
The same limitation of Deep Neural Networks (DNNs) in image classification is also noticed if the input is affected by affine transformations that do not modify the pixels but their relative position in space.
The most common means of limiting these problems is to increase the generalization level of a CNN, which is achievable using different methods. Some research works proposed to increase the depth of CNN architectures~\cite{ResNet20}, others proposed to modify the hyper-parameters~\cite{Young2015hyperparameters} and using data pre-processing during the training~\cite{Mikolajczyk2018dataaugmentation}.
For a CNN, the convolutional and the Max Pooling layers provide the generalization and the capability to detect high-order features in a large region of the image (\textit{invariance property}), but without preserving any relation with other identified features.

With the introduction of the Capsule Networks (CapsNets) by Google~\cite{Transforming_autoencoders}, the basic building block of a neural network, i.e., the neuron, has been replaced by a group of neurons, called \textit{capsule}. 
The capsules encode spatial information in a vector form. When a detected feature moves around the image, the probability of being detected does not vary, but its \textit{pose} information changes (\textit{equivariance property}). 
The work in~\cite{Sabour} proposes an efficient way of learning the coupling between capsules from different layers through the so-called \textit{dynamic routing} algorithm, an iterative process that replaces the behavior of the max pooling, but without losing any information. Hence, such a capsule structure improves the network's generalization because it can efficiently learn cross-correlations between different features of the inputs. Recently, Rajasegaran et al.~\cite{DeepCaps_keras} showed that a deeper version of CapsNets can achieve high accuracy also on mid-complex datasets like the CIFAR10~\cite{CIFAR}, despite reducing the number of parameters compared to the shallower CapsNet in~\cite{Sabour}.

Existing works~\cite{Improving_the_Robustness}\cite{Marchisio_2023MICPRO_SeVuC}\cite{DBLP:Michels}\cite{Marchisio_2022IJCNN_fakeWeather} have analyzed the vulnerabilities and robustness of CapsNets against affine transformations and adversarial attacks, respectively. \textit{However, they lack a systematic study comparing different types of CapsNets and CNNs and a detailed analysis of the impact of different CapsNet functions (like dynamic routing) on the robustness.} 
Moreover, Michels et al.~\cite{DBLP:Michels} did not investigate the CapsNets' robustness when an adversarial defense, such as the adversarial training~\cite{Madry}, is applied.

Such analyses would establish an understanding of differences between CNNs and CapsNets w.r.t. the robustness against adversarial attacks and how the robustness of CapsNets changes depending on the model features. This could help future CapsNet designs in \textit{accounting for the security vulnerabilities into design constraints}, increasing the applicability of CapsNets in real-world scenarios~\cite{Shafique_2021ICCAD_EnergyEfficientSecureEdgeAI}.

\vspace{3pt}

\textbf{Research Questions and Associated Challenges}\\
The goal of our paper is to investigate these research questions:

\begin{enumerate} 
 \item \textit{Are CapsNets more robust than CNNs against adversarial attacks and affine transformations?}
 \item \textit{If yes, how can these phenomena be analyzed in a systematic way?}
 \item \textit{Which CapsNet functions contribute more to the robustness improvement?}
\end{enumerate}

Answering these questions is a challenging task. Firstly, we evaluate a good metric of comparison between CapsNets and CNNs, i.e., which network models give a fair and significant robustness comparison, which types of adversarial attacks are applied, etc. Then, it should be interesting to analyze the transferability of the adversarial attacks, i.e., white-box attacks. \textit{If an adversarial example has been generated to fool network A, does it also fool network B?}

\begin{figure}[t]
 \centering
 \includegraphics[width=\linewidth]{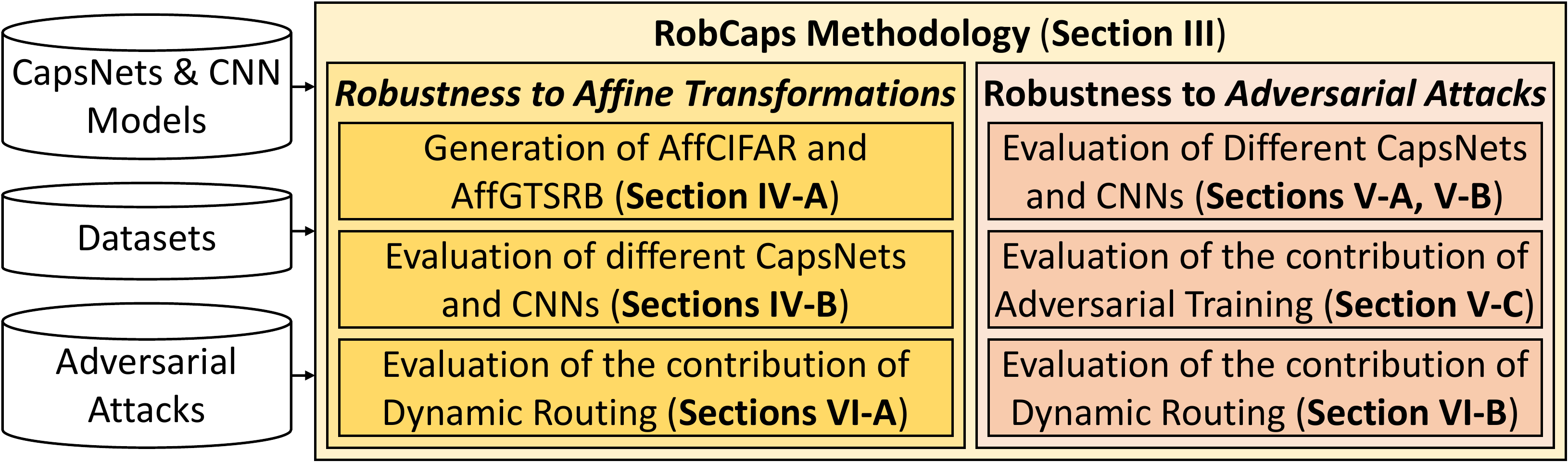}
 \caption{Overview of our novel contributions in this work.}
 \label{fig:Novel_contributions}
\end{figure}

\vspace{3pt}
\textbf{Our Novel Contributions are} (see \Cref{fig:Novel_contributions}):

\begin{itemize}
 \item We generate an affined-transformed version of the CIFAR10 and GTSRB datasets, called \textbf{affCIFAR} and \textbf{affGTSRB}, respectively. \textit{(Section~\ref{subsec:affine_cifar10})}
 \item We evaluate / compare the \textbf{robustness of different CapsNets and CNNs (like ShallowCaps, DeepCaps, ResNet20) against affine trans-formations} for different datasets and different networks. \textit{(Section~\ref{subsec:affine_transformations_results})}
 \item We compare the robustness of different networks \textbf{against adversarial attacks} for different datasets. Further analyses have been carried out in the presence of a defense such as the \textbf{adversarial training}. \textit{(Section~\ref{sec:results_attacks})}
 \item We evaluate the role of the \textbf{dynamic routing} towards the CapsNets robustness. \textit{(Section~\ref{sec:robustness_routing})}
\end{itemize}

In summary, our key results depicted in \Cref{tab:summary_results} show that the DeepCaps~\cite{DeepCaps_keras} is more robust than a deeper ResNet20~\cite{ResNet20} against affine transformation and different types of adversarial attacks, increasing the complexity of the input data. As we will demonstrate, such improvements in the robustness also hold when the adversarial examples are transferred from one network to the other and vice-versa.

\begin{table}[t]
\centering
\caption{Key results obtained in this paper.}
\resizebox{\linewidth}{!}{\begin{tabular}{c|c|c|c}
\hline \hline
\multicolumn{2}{c|}{ } & DeepCaps & ResNet20 \\ \hline
Affine & Accuracy AffNIST & $87.60\%$ & $\textbf{96.39\%}$ \\
Transformations & Accuracy AffGTSRB & $81.14\%$ & $\textbf{89.75\%}$ \\
Robustness & Accuracy AffCIFAR & $\textbf{78.66\%}$ & $75.84\%$ \\ \hline \hline 
Adversarial & Accuracy MNIST PGD, $\varepsilon = 0.05$ & $1.20\%$ & $\textbf{86.78\%}$ \\
Attacks & Accuracy GTSRB PGD, $\varepsilon = 0.02$ & $\textbf{50.35\%}$ & $18.99\%$ \\
Robustness & Accuracy CIFAR10 PGD, $\varepsilon = 0.005$ & $\textbf{37.49\%}$ & $1.41\%$ \\ \hline \hline
\end{tabular}
}
\label{tab:summary_results}
\end{table}

After showing the power of the capsules, we focus our analysis on the dynamic routing, which increases the confidence of the prediction, with a consequent improvement in terms of accuracy. By knowing that, our challenging question is: \textit{Is the dynamic routing also helpful in guaranteeing the CapsNets robustness?}
Our results and analyses provide great insights when relating CNNs and CapsNets against adversarial attacks and affine transformations, as well as how CapsNets' behavior changes when modifying model features.

Before proceeding to the technical sections, we discuss adversarial attacks and CapsNets in Section~\ref{sec:background} to a level of detail necessary to understand our contributions.

\section{Background and Related Work}
\label{sec:background}

\subsection{Adversarial Attacks}

\begin{figure}[t]
 \centering
 \includegraphics[width=.85\linewidth]{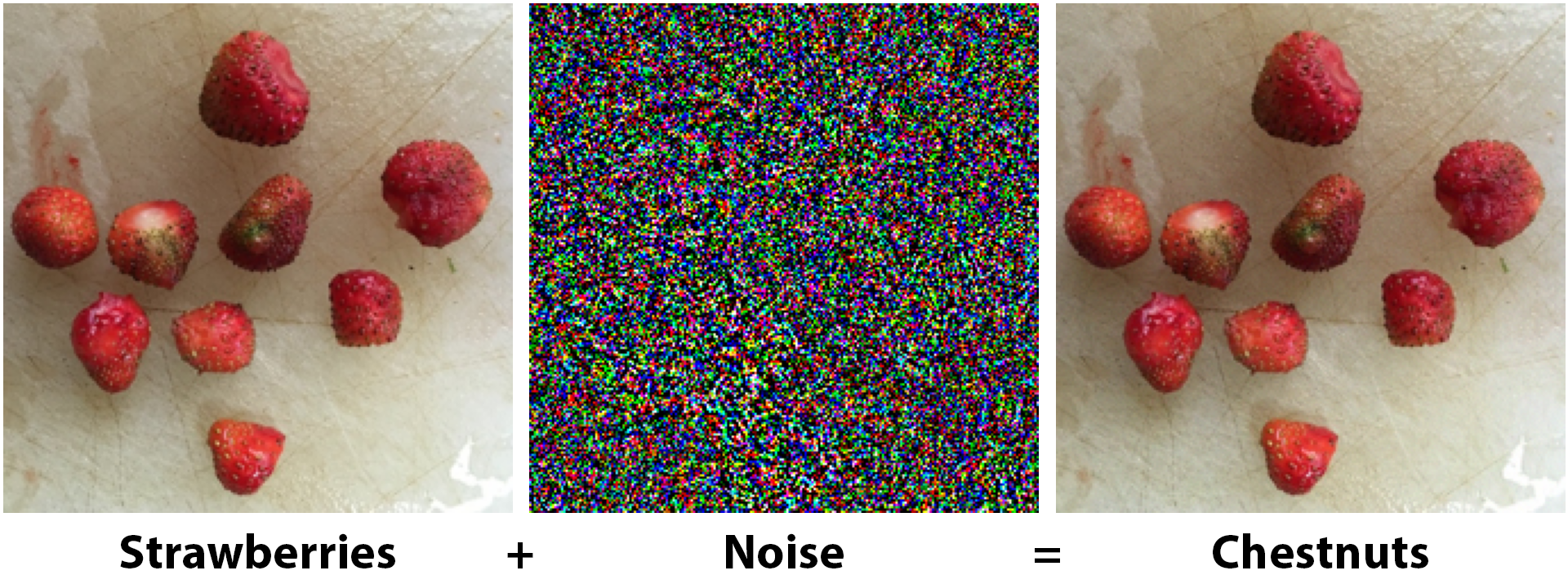}
 \caption{Example of an adversarial attack's functionality, where strawberries are misclassified as chesnuts~\cite{Goodfellow}.}
 \label{fig:example_attack_goodfellow}
\end{figure}

 Formally, having an example $x$ that is correctly classified by a well-trained model $M(x) = y_{true}$, an adversarial example $x' = x + \eta$ is defined as a new input, perceptually identical to the original one, but wrongly classified by the model, i.e., $M(x') \neq y_{true}$. 
%
%
Goodfellow et al.~\cite{Goodfellow} proposed the fast gradient sign method (FGSM), a white-box attack to generate adversarial examples by exploiting the gradient of the model w.r.t. the input image, towards the direction of the highest loss. An example of its functionality is shown in \Cref{fig:example_attack_goodfellow}, where the crafted noise added to the original input is imperceptible to the human eye but results in a misclassification. Afterwards, Madry et al.~\cite{Madry} and Kurakin et al.~\cite{Kurakin} proposed two different versions of the projected gradient descent (PGD) attack, which is an iterative version of the FGSM that introduces a perturbation $\alpha$ to multiple smaller steps. After each iteration, the PGD projects the generated image into a ball with a radius $\varepsilon$, keeping small the size of the perturbation. 
It is a white-box attack and has both the targeted and untargeted versions. The algorithm consists of the iteration expressed in \Cref{eq:PGD}, where $\theta$ is the set of parameters and $t$ is the target label.

\vspace*{-5pt}

\begin{equation}
 x'_{i} = x'_{i-1} - proj_{\varepsilon} (\alpha\cdot sign(\nabla_x loss(\theta,x,t)))
 \label{eq:PGD}
\end{equation}

\begin{figure*}[t]
 \centering
 \includegraphics[width=.75\textwidth]{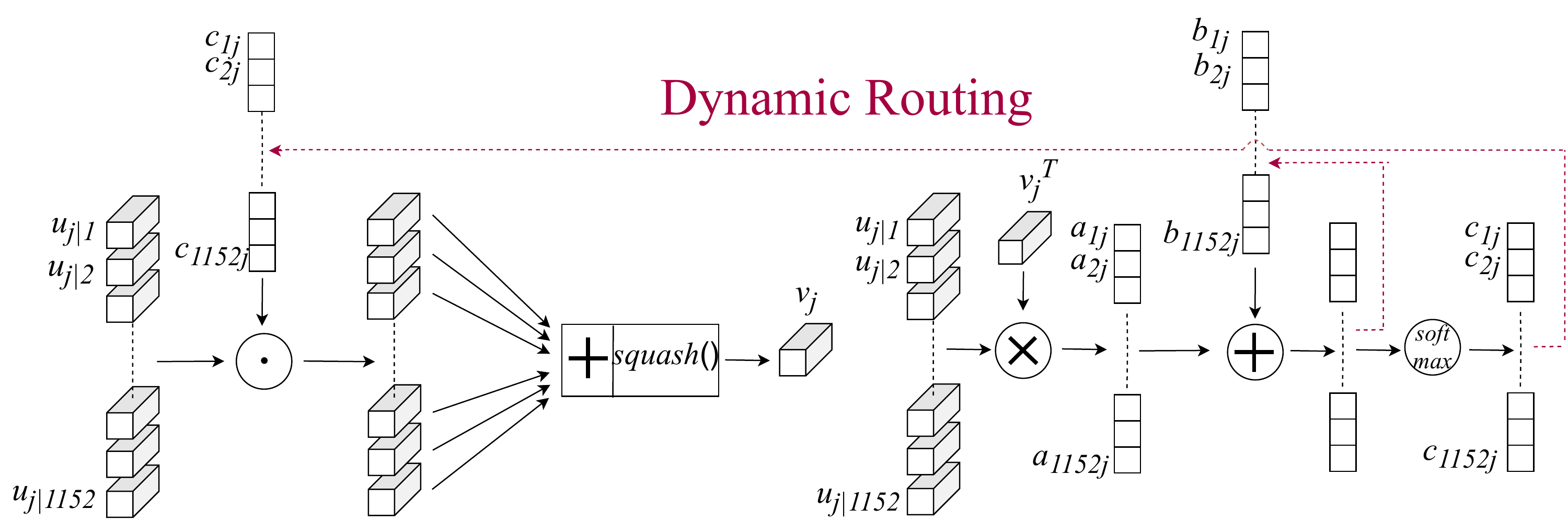}
 \caption{Schematic overview of the processing flow occurring in the Dynamic routing of the DigitCaps layer.}
 \label{Dynamic_Routing}
\end{figure*}


Carlini and Wagner~\cite{Carlini_Wagner} proposed a powerful white-box targeted attack method that exploits $l_{\infty}$, $l_1$ and $l_2$ distances to preserve the imperceptibility of the adversarial example.
It is performed by solving the optimization problem expressed in \Cref{eq:CW_attack}.

\vspace*{-5pt}

\begin{equation}
 ||{\eta}||_2 + c \cdot max(G(x,\eta,t) - M(x,\theta)_t, -k)
 \label{eq:CW_attack}
\end{equation}
 
The algorithm aims to minimize both the components of the equation: (i) the distance $\eta$ between the input and the adversarial image and (ii) the distance between the max output activation ($G(x,\eta,t):= max_{i\neq t}(M(x + \eta))$) and the confidence $M(x)_t$ of the target label $t$. 
The value $c$ is updated at every iteration to balance the two terms during the generation of the attacked data.
Many works showed the success of such attacks in fooling DNNs and provided state-of-the-art success rate results~\cite{Goodfellow}\cite{Carlini_Wagner}\cite{Madry}. 
A common countermeasure to defend against such attacks is the adversarial training~\cite{Madry}, which extends the training set for DNNs by also including the adversarial examples.


\subsection{Capsule Networks}


CapsNets gathered attention due to their capability to achieve higher classification accuracy than traditional CNNs.
Sabour et al.~\cite{Sabour} introduced the first CapsNet architecture, based on the following differences w.r.t. traditional CNNs:

\begin{itemize} 
 \item \textit{capsules}: multi-D entities, instead of single neurons, that constitute each layer.
 \item a \textit{dynamic routing} algorithm between two adjacent capsules selects the capsules that must be propagated, based on their pose agreement.
 \item a \textit{squash} function compresses the components of each capsule in a small interval at the end of each layer.
\end{itemize}

The architecture designed in~\cite{Sabour}, which we call \textit{ShallowCaps} (for ease of discussion), is composed of:

\begin{itemize} 
 \item a first standard convolutional layer with $256$ $9 \times 9$ kernels.
 \item a Primary Capsule layer, convolutional with $9 \times 9$ kernels and the same parameters as the previous layer, but reshaped to form $32$ $8$-dimensional capsules.
 \item a DigitCaps layer of $10$ capsules of dimension $16$.
\end{itemize}

The last layer defines a transformation matrix that, during the training, learns the relationship between all the capsules of the Primary Capsule layer and the capsules of the DigitCaps layer. 
The dynamic routing (Fig.~\ref{Dynamic_Routing}) has the task of propagating only the activations with a high contribution by updating a set of coupling coefficients. Specifically, this iterative algorithm ensures that only the most voted opinion among the predictions is propagated to the DigitCaps layer.

\noindent
 The limit of this architecture is that it cannot correctly generalize a complex dataset like the CIFAR10. Kumar et al.~\cite{Kumar} proposed a three-layer architecture, like the previous one, for the GTSRB dataset~\cite{GTRSB}, increasing the number of capsules coupled with the DigitCaps layer. This one needs a huge number of parameters and wasteful use of resources to reach similar performances as traditional CNN models. 
To solve this problem, the DeepCaps~\cite{DeepCaps_keras} has been designed to reduce the number of parameters, exploiting deeper capsule architectures.
Without stacking more than one fully-connected layer of capsules, the DeepCaps introduces a new kind of 3D dynamic routing that exploits 3D convolutions.

Both the dynamic routing and the expectation-maximization routing used by Hinton et al.~\cite{matrix_capsules} are computationally expensive in terms of execution time. Many works tried to accelerate the procedure at the algorithmic level~\cite{Zao}\cite{Marchisio_2020IJCNN_FasTrCaps}\cite{Marchisio_2020ICCAD_NASCaps} or at the hardware level~\cite{Marchisio_2019DATE_CapsAcc}\cite{Marchisio_2020DAC_QCapsNets}\cite{Marchisio_2020DATE_ReDCaNe}\cite{Marchisio_2022ISLPED_CapsNetApproximateSquashSoftmax}\cite{Marchisio_2021TVLSI_FEECA}\cite{Marchisio_2021TCAD_DESCNet}, and others proposed novel routing strategies~\cite{attention_routing}\cite{Li}.
On the contrary, many other works proposed to incorporate the routing procedures into the training process, removing it. In other words, it is possible to learn the coupling coefficients implicitly, including them in the weights of the transformation matrix. Furthermore,~\cite{self_routing} proposed a different algorithm introducing new coupling weights between two capsule layers, called \textit{self-routing}.

\begin{figure*}[t]
 \centering
 \includegraphics[width=.75\linewidth]{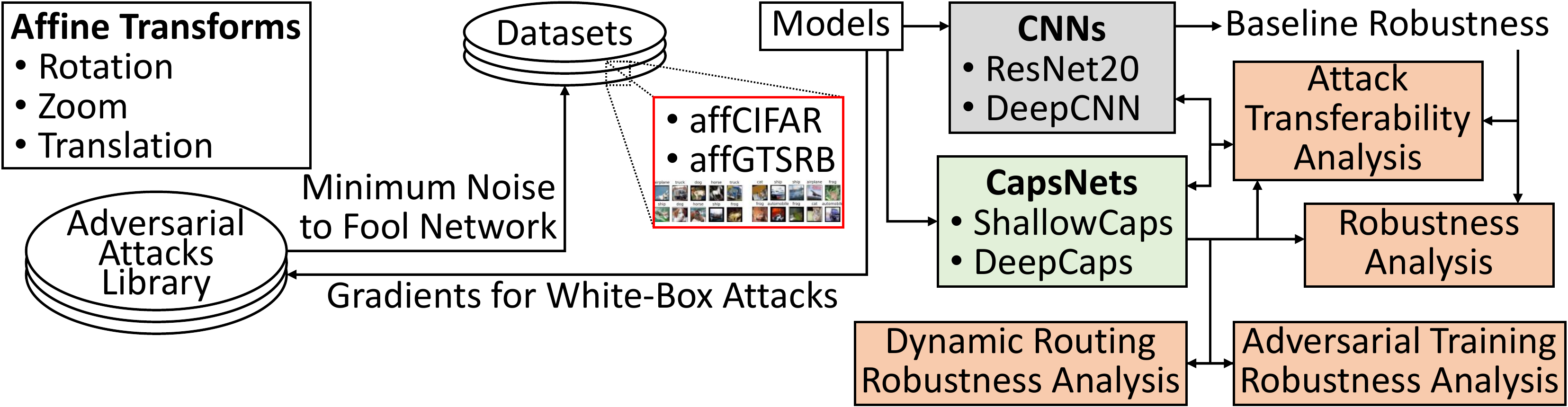}
 \caption{Overview of our RobCaps methodology.}
 \label{fig:methodology}
\end{figure*}

Our analysis (\textit{Section~\ref{sec:robustness_routing}}) also proves that the contribution of the dynamic routing against attacks and affine transformations is not effective. Then, incorporating it into the training process could be a solution to avoid this expensive procedure.

Recent works showed the vulnerability of CapsNet against adversarial attacks.
Frosst et al.~\cite{Frosst} investigated the detection of adversarial examples using the reconstruction quality of the CapsNets. Peer et al.~\cite{Peer} and Marchisio et al.~\cite{Marchisio} applied the FGSM method~\cite{Goodfellow} and their proposed attack on CapsNets, respectively. Michels et al.~\cite{DBLP:Michels} compared the results of different attacks on CapsNets trained on different datasets. The RoHNAS framework~\cite{Marchisio_2022Access_RoHNAS} includes adversarial robustness among the optimization objectives and conducts Neural Architecture Search to obtain energy efficient and robust CapsNets.
However, before employing CapsNets in safety-critical applications, their robustness must be analyzed in practical use-case scenarios, e.g., investigating applications where the CapsNets' classification accuracy is on par or better than the state-of-the-art DNNs, and when robust defenses like adversarial training are adopted.

\section{In-Depth View of our RobCaps Methodology}
\label{sec:methodology}

The CapsNets has been considered relatively more robust towards adversarial attacks when compared to traditional CNNs.
To investigate this intuition, we present a detailed analysis to answer our main research questions, and to show (1) if and why the Capsule Network under attack provides a better response than traditional CNNs, (2) which model quality plays an important role, and their limits.
Knowing the main differences between CapsNets and traditional CNNs, we explore the impact of these networks on affine transformations and adversarial attacks. 
Moreover, we study the role of different functions of a CapsNet on the robustness against these attacks. 
Towards a fair and comprehensive evaluation, the results for the ShallowCaps have been compared with three different architectures (chosen according to their properties, their number of parameters, and their depth) for three different datasets, i.e., MNIST~\cite{mnist}, GTSRB~\cite{GTRSB} and CIFAR10~\cite{CIFAR}.

\begin{itemize} 
\item A deeper CapsNet architecture, like the \textit{DeepCaps} model~\cite{DeepCaps_keras}. Despite being deeper than the ShallowCaps, it has fewer parameters.
The \textit{DeepCaps} employs four groups of 2D convolutional capsule layers, with a 3D convolution layer in the last group and a fully connected capsule layer of $10$ 32D capsules.
\item \textit{ResNet20} (He et al.~\cite{ResNet20}) is one of the best performing CNN architectures for CIFAR10, used in various applications. It would be interesting to compare the capabilities of the CapsNet with a widely used CNN, which is deeper and employs Residual Blocks with convolutional and average pooling layers.
\item A traditional \textit{CNN} with the same depth as the DeepCaps, but without multidimensional entities such as capsules. The dimensions of the layers are reshaped in a 2D fashion, using traditional convolutional layers with batch normalization instead of capsules with squash compression, and a traditional fully connected layer instead of the DigitCaps layer with dynamic routing.
Its comparison w.r.t the DeepCaps highlights the contribution to the robustness of 3D convolutions and capsules.
\end{itemize}

\subsection{Step-By-Step View of our Methodology}

Our methodology, shown in Fig.~\ref{fig:methodology}, is composed of these following steps:

\begin{enumerate}[label=\arabic*)]
\item \textbf{Evaluation of robustness on affine transformations:}

\begin{enumerate}[label=\roman*),labelwidth=1pt]

\item Train our networks with the clean datasets using the same hyperparameters and data augmentation.
\item Generate the affine-transformed version of each dataset for a given set of affine-transformations. For the CIFAR10 and the GTSRB datasets, we design two novel transformed datasets with random translations, rotations, and zooms (which we call \textit{affCIFAR} and \textit{affGTSRB}, see Section~\ref{subsec:affine_cifar10}).
\item Use such affCIFAR and affGTSRB datasets for inference, as the case for the already existing affNIST~\cite{affNIST}, to evaluate the response of the networks to affine transformations. 
\end{enumerate}

\item \textbf{Evaluation of robustness on adversarial attacks:}
\vspace*{3pt}

We use the saved parameters of the trained models to evaluate the gradient, with respect to the input, for the two implemented white-box attacks. The key steps of our methodology are:
\begin{enumerate}[label=\roman*),labelwidth=1pt]
\item Apply the projected gradient descent (PGD) attack for each architecture and dataset.
\item Test the networks with the generated adversarial inputs, evaluating the accuracy behavior, increasing the perturbation level. 
\item Apply the Carlini Wagner attack (CW) for each dataset.
\item Evaluate the mean distortion required by the algorithm to misclassify $500$ images of the test datasets and its fooling rate.
\item Apply at the input to a network the adversarial image generated with another one to test the transferability of the attack. 
\item Test the robustness when the adversarial training defense is applied.
\end{enumerate}

\item \textbf{Analyzing the contribution of the dynamic routing to the CapsNet's robustness:}

\begin{enumerate}[label=\roman*),labelwidth=1pt]
 \item Modify the dynamic routing of the DigitCaps layer of the DeepCaps and then generate three versions of it with different routing algorithms.
 \item Analyze the robustness against affine transformations.
 \item Analyze the robustness against PGD and CW attacks.

\end{enumerate}
\end{enumerate}

\begin{figure*}[t]
 \centering
 \includegraphics[width=.7\textwidth]{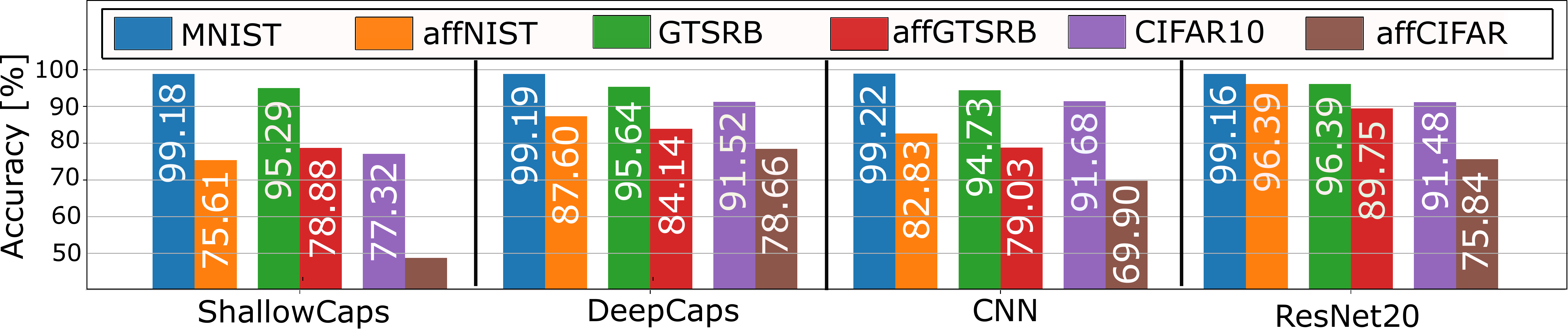}
 \caption{Robustness against affine transformations. }
 \label{affine_result}
\end{figure*}

\subsection{Experimental Setup}

These architectures have been trained with the $40 \times 40$ sized version of the MNIST dataset and tested on the affNIST for evaluating the robustness against affine transformations.
For all the architectures tested on CIFAR10, input data have been resized before the training, from $32 \times 32$ to $64 \times 64$, following the pre-processing steps used in~\cite{DeepCaps_keras}. For the GTSRB dataset, the input images' size is kept at $32 \times 32$.
The data augmentation and hyperparameters used for the training are kept the same for all the networks.
As a regularization term, the CapsNets have the reconstruction loss provided by the decoder. For the evaluation of the loss, we use the same function as in~\cite{Sabour} for CapsNets and the Cross-Entropy for CNNs.

We implemented the attack algorithms using the CleverHans~\cite{cleverhans} library, adapted for the Keras framework~\cite{keras} with Tensorflow backend~\cite{tensorflow}. 
The networks have been trained on multiple Nvidia RTX-2080Ti GPUs with CUDA~$10$. 
To have a good comparison metric, we train different versions of the DeepCaps architecture modifying/removing the dynamic routing.

\section{Robustness Against Affine Trasformations}
\label{sec:results_affine_trasformations}

\subsection{Affine-CIFAR10 (affCIFAR) and Affine-GTSRB (affGTSRB) Datasets Generation}
\label{subsec:affine_cifar10}

While a dataset with affine transformed images of the MNIST dataset (affNIST) is already available, we create an affine version of the CIFAR10 and GTSRB datasets, which we call \textit{affCIFAR} and \textit{affGTSRB}, to compare the response of the networks defined in Section~\ref{sec:methodology}.
The test data was created by modifying the $10\,000$ test images from the original dataset with random affine transformations.
Every image is transformed following these criteria:

\begin{itemize} 
 \item \textit{Translations:} random pixels translations in one or in two dimensions by a factor between $10\%$ and $25\%$ of the input image size, with a fixed interval.
 \item \textit{Rotations:} random rotations between $+20$ and $-20$ degrees with a fixed step.
 \item \textit{Zooms:} the vertical and horizontal expansions are chosen uniformly between $0.8$ (i.e., shrinking the image by $20\%$) and $1.2$ (i.e., enlarging the image by $20\%$).
\end{itemize}

\begin{figure*}[t]
 \centering
 \includegraphics[width=.8\linewidth]{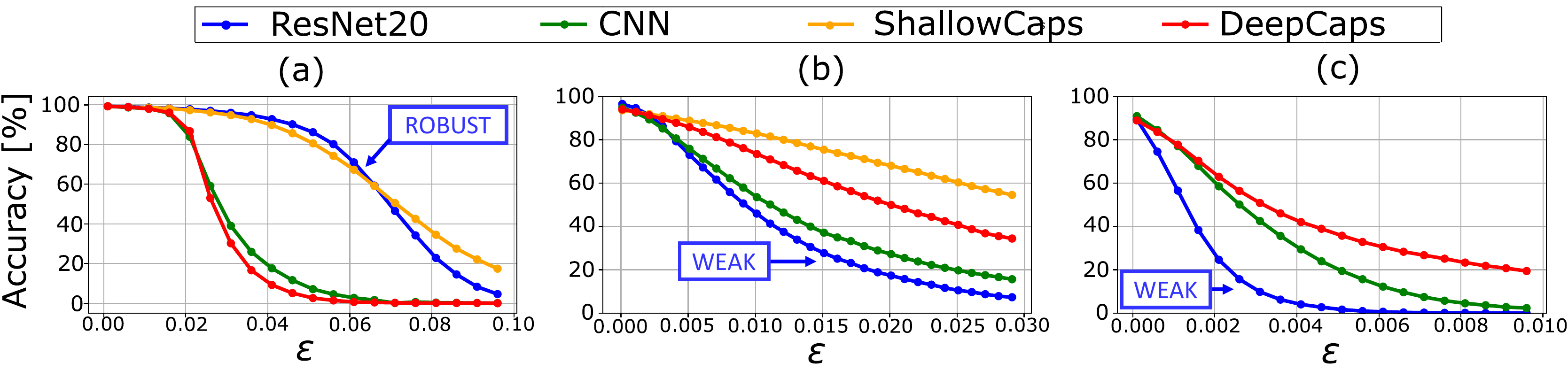}
\caption{Robustness against the PGD attack for (a) the MNIST, (b) the GTSRB, and (c) the CIFAR10 datasets. }
 \label{PGD}
\end{figure*}

\subsection{Affine Trasformations Results}
\label{Affine trasformations}
\label{subsec:affine_transformations_results}

For each model defined in Section~\ref{sec:methodology}, we evaluate the accuracy for all the datasets and their respective affine-transformed versions. The results are shown in Figure~\ref{affine_result}.
\vspace*{3pt}

\noindent
\textbf{ShallowCaps vs. DeepCaps}: 
As shown in Figure~\ref{affine_result}, the ShallowCaps on the CIFAR10 dataset achieves lower accuracy than the state-of-the-art ($77.32\%$). Such limitation is solved by the DeepCaps, which reaches better results even when using the affine version of the respective dataset ($78.66\%$). 
Thus, using a deeper architecture while keeping the same capsule structure, the DeepCaps model has fewer parameters while having better generalization. Its accuracy with the CIFAR10 dataset ($91.52\%$) and with the affine transformed datasets are much higher compared to ShallowCaps.
In fact, despite the shallower model reaching a good accuracy on the normal MNIST and GTSRB datasets, it is still unable to generalize as the DeepCaps against affine transformations.
The improvement could also be explained by the presence of the 3D convolutional layer.
The effect of having 3D convolutions, compared to a stack of fully connected capsules, is similar to when we compare the generalization level offered by the Multi-Layer Perceptrons (MLP) and the CNNs.
In the DigitCaps layer, each element of the transformation matrix learns if a capsule is correlated with each capsule of the following layer. 
On the contrary, with the 3D convolution, sliding a 3D kernel, the same weights are used among all the capsules of the layer. This characteristic also allows learning the presence of a particular feature if the input image is spatially transformed (e.g., translated, rotated, or zoomed), preserving the capsule structure. 
\vspace*{3pt}

\noindent
\textbf{DeepCaps vs. CNN and ResNet20}:
Another significant result is provided by comparing the response of the DeepCaps with a traditional CNN, having a similar base architecture. While the accuracy of the CNN on the MNIST, GTSRB, and CIFAR10 datasets is similar to the DeepCaps, the CNN's robustness against the affNIST, affGTSRB, and affCIFAR is much lower. These results confirm the benefits of capsules against affine transformations.
Compared to the DeepCaps, the ResNet20 is deeper but has fewer parameters. It can generalize better for the affMNIST and affGTSRB but worse for the affCIFAR dataset. This apparently contradictory result is due to the difference in complexity between the datasets. While for simple datasets, a deep CNN, like the ResNet20, can generalize very well, for more complex tasks like the affCIFAR, it is outperformed by the DeepCaps. This observation highlights the capability of the capsule architectures to preserve spatial correlations between the features detected, and this difference w.r.t deeper traditional CNNs is even more evident when the input dataset is composed of complex features like the CIFAR10.

\begin{figure*}[t]
 \centering
 \includegraphics[width=.85\textwidth]{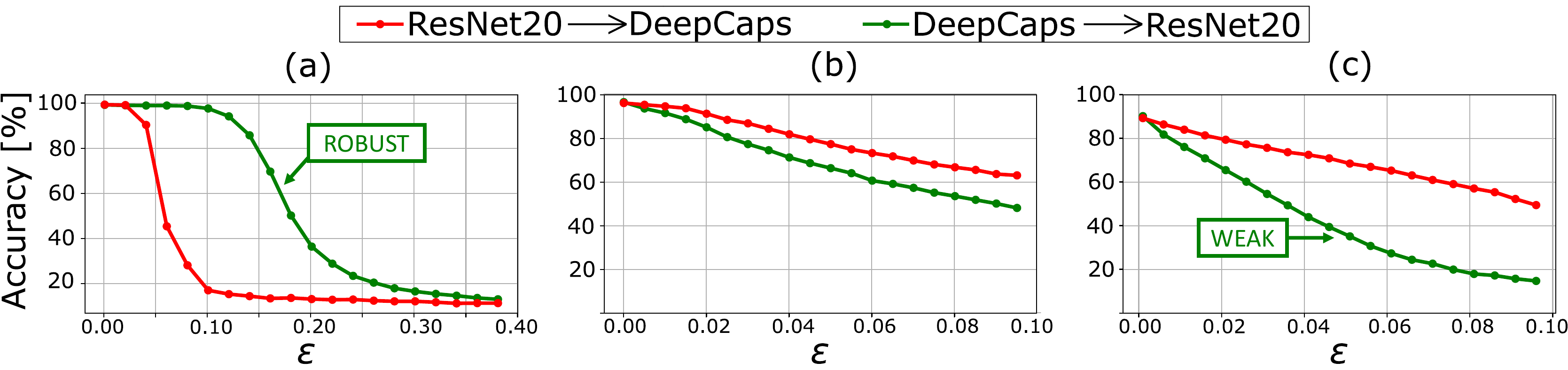}
 \caption{Transferability for the PGD attack: comparison of the network response for (a) MNIST, (b) GTSRB, and (c) CIFAR10 datasets.}
 \label{PGD_TRANSFERABILITY}
\end{figure*}

\section{Robustness Against Adversarial Attacks}
\label{sec:results_attacks}

\subsection{Evaluations under the PGD Attack}

We analyze the network response by increasing the perturbation level $\varepsilon$, generated by the algorithm.
Figures~\ref{PGD}a,~\ref{PGD}b, and~\ref{PGD}c show the results for the MNIST, GTSRB, and CIFAR10 datasets, respectively.

\vspace*{3pt}
\noindent
\textbf{ShallowCaps vs. ResNet20}:
Applying the PDG attack for the MNIST dataset, the ResNet20 is less vulnerable than other networks for low levels of $\varepsilon$.
The ShallowCaps robustness behavior, not so far from the one of the ResNet20, overperforms the ResNet20 when $\varepsilon \approx 0.065$. 
Hence, despite the low number of layers, the ShallowCaps responds to the PGD attack similarly to a deeper CNN. 
\vspace*{3pt}

\noindent
\textbf{DeepCaps vs. ShallowCaps}: According to the results, the ShallowCaps is more robust than the DeepCaps, in contrast to what happens for affine transformations. This means that increasing the depth of a CapsNet does not provide more robustness against perturbed images.
Note that the ShallowCaps response for the CIFAR10 dataset has not been examined because of its very low baseline accuracy, which is not comparable with other networks. 
\vspace*{3pt}

\noindent
\textbf{DeepCaps vs. ResNet20 vs. CNN}:
For this kind of algorithm and the MNIST dataset, Figure~\ref{PGD}a shows that the DeepCaps behaves worse than the ResNet20.
On the contrary, for more complex datasets like CIFAR10 or GTSRB, the results in Figure~\ref{PGD} show that the ResNet20 is not as robust as for the MNIST dataset.
By increasing the perturbation's size, the attack's success rate grows faster than on DeepCaps. Such an outcome is in line with the takeaway from Section~\ref{Affine trasformations}, which showed the DeepCaps be more robust than the ResNet20 against the transformations in affCIFAR. 

The behavior of the CNN curve for GTSRB and CIFAR10 always stays below the curve of the DeepCaps. It means that the capsule architecture plays a fundamental role in improving the robustness against the PGD attacks when the dataset becomes more complex than the MNIST.
\vspace*{3pt}

\noindent
\textbf{Transferability ResNet20 $\Longleftrightarrow$ DeepCaps:}
Towards a more comprehensive study of the robustness against the PGD, we analyze the transferability of the attacks, between the ResNet20 and the DeepCaps, presenting the two opposite behaviors. We provide as inputs to the DeepCaps the adversarial examples generated with the gradient of the ResNet20 and vice-versa. Figure~\ref{PGD_TRANSFERABILITY} shows the transferability between these two networks for different datasets.

For the MNIST dataset, the attacks generated for the ResNet20, tested on DeepCaps, have a more significant effect than the opposite way. As shown in Figure~\ref{PGD_TRANSFERABILITY}a, this outcome confirms that the ResNet20 appears suitable for the generalization of the MNIST. The contrasting results can be derived for the GTSRB and CIFAR10 dataset, where the DeepCaps shows greater robustness than the ResNet20 due to a better generalization ability for a more complex dataset.
\vspace*{-3pt}

\begin{table*}[t]
\centering
\caption{Robustness results against the CW attack.}
\resizebox{.85\linewidth}{!}{\begin{tabular}{c|cc|cc|cc}\hline\hline
 & \multicolumn{2}{c|}{MNIST} & \multicolumn{2}{c|}{GTSRB} & \multicolumn{2}{c}{CIFAR10} \\ 
Network & Mean Distortion & Fooling Rate & Mean Distortion & Fooling Rate & Mean Distortion & Fooling Rate\\\hline\hline
ShallowCaps & $1.59$ & $98.6\%$ & $1.31$ & $100\%$ & - & - \\
Deepcaps & $1.24$ & $86.8\%$ & $1.16$ & $98.8\%$ & $0.34$ & $100\%$ \\
CNN & $0.95$ & $100\%$ & $0.59$ & $100\%$ & $0.23$ & $100\%$ \\
ResNet20 & $0.94$ & $100\%$ & $0.34$ & $100\%$ & $0.12$ & $100\%$ \\ \hline \hline 
\end{tabular}
}
\label{Carlinin_Wagner_results}
\end{table*}

\begin{figure*}[t]
 \centering
 \includegraphics[width=.9\textwidth]{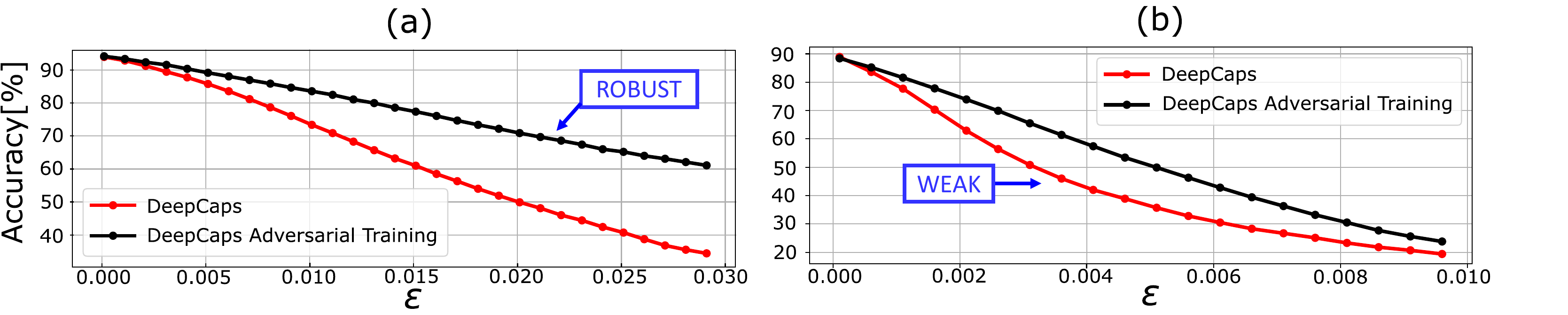}
 \caption{Adversarially vs. normally trained DeepCaps with (a) the GTSRB and (b) the CIFAR10 datasets.}
 \label{PGD_ADVERSARIAL_TRAINING}
\end{figure*}


\subsection{Evaluations under the Carlini Wagner (CW) Attack}

To have a more solid comparison, the CapsNets and CNNs have also been tested against the CW attack, a different kind of algorithm that does not define a threshold for the magnitude of the perturbation (like the $\varepsilon$ in the PGD attack).
It is an iterative targeted algorithm that tries to reduce the gap between the target and the predicted class (success rate) with the minimum perturbation (mean distortion), estimated as the $l_{2}$ distance. 
For a more robust network, the algorithm necessitates more iterations to obtain that the probability of the target class overcomes the probability of the correct class. 
As a consequence, more iterations also imply a higher $l_{2}$ distance between the original image and the adversarial example.
For our estimations, we set a maximum of $10$ iterations for the MNIST and $5$ iterations for the CIFAR10 dataset. In addition, for the attacks on CIFAR10, the algorithm has been forced to set the confidence of the targeted class to $0.5$ higher than the confidence of the true label.
Table~\ref{Carlinin_Wagner_results} reports the fooling rate 
and the mean distortion for both the datasets.
\vspace*{3pt}

\noindent
\textbf{CapsNets vs. CNNs:} The CW attack is very effective for traditional CNNs. In fact, it reaches a $100\%$ fooling rate for all three datasets. Similar findings were also made in~\cite{Carlini_Wagner}. On the other hand, both CapsNets show greater robustness (i.e., lower fooling rate) than CNNs, for the MNIST dataset (and also for the GTSRB, even if the fooling rate of the DeepCaps is just a little bit lower than $100\%$). The CapsNets also require a higher mean distortion than the CNNs, which makes the resulting adversarial example more perceptible. For the CIFAR10 dataset, the CW attack shows its effectiveness because, for all the networks, the fooling rate is $100\%$. However, we can notice that CapsNets are more robust due to a higher mean distortion. 

\noindent
\textbf{DeepCaps vs. ShallowCaps:}
The DeepCaps appears to be more robust than the ShallowCaps, because of a lower fooling rate, despite having slightly lower mean distortion. Therefore, the depth and the 3D convolutions help to generalize better against the CW attack. 
\vspace*{3pt}

\noindent
\textbf{Transferability ResNet20 $\Longleftrightarrow$ DeepCaps:}
Table~\ref{table_accuracies_results_transferability} shows the transferability of the attacks between ResNet20 and DeepCaps for the CW attack. The values report the accuracies of the two models that receive as input a sample of $500$ targeted adversarial examples generated by the CW algorithm applied to the other network. The high accuracy values demonstrate the low level of transferability of the CW attack. Despite this, the ResNet20 still achieves lower accuracies than the DeepCaps, thereby performing less robustly.

 \begin{table}[t]
 \caption{Transferability of the CW attack between the DeepCaps and the ResNet20.}
 \begin{center}
 \begin{tabular}{cccc} \hline \hline
 	Network & MNIST & GTSRB & CIFAR10 \\ \hline \hline
 	DeepCaps $\rightarrow$ ResNet20 & $97.4\%$ & $94.0\%$ & $86.8\%$ \\
 	ResNet20 $\rightarrow$ DeepCaps & \textbf{97.8\%} & \textbf{95.4\%} & \textbf{89.2\%} \\
 \hline
 \end{tabular}
 \end{center}
 \label{table_accuracies_results_transferability}
\end{table}

\subsection{DeepCaps defended with the PGD Adversarial Training}

\begin{table*}[t]
\centering
\caption{Adversarially and normally trained DeepCaps against the CW attack.}
\begin{center}
\resizebox{.7\linewidth}{!}{\begin{tabular}{c|cc|cc}\hline\hline
 & \multicolumn{2}{c|}{GTSRB} & \multicolumn{2}{c}{CIFAR10} \\ 
Network & Mean Distortion & Fooling Rate & Mean Distortion & Fooling Rate\\\hline\hline
Normally trained DeepCaps & $1.16$ & $98.8\%$ & $0.34$ & $100\%$ \\
Adversarially trained DeepCaps & \textbf{1.44} & \textbf{98.6\%} & \textbf{0.84} & \textbf{96.6\%} \\\hline\hline 
\end{tabular}
}
\end{center}
\label{tab:Carlinin_Wagner_adversarial_training}
\end{table*}

\begin{table*}[t!]
\caption{Robustness results against affine transformations.}
 \begin{center}
 \resizebox{.8\linewidth}{!}{\begin{tabular}{c|ccc|ccc} \hline \hline
 	Network & MNIST40 & GTSRB & CIFAR10 & AffNIST & AffGTSRB & AffCIFAR \\ \hline \hline
 	DeepCaps without dynamic routing & \textbf{99.27\%} & \textbf{96.27\%} & $91.47\%$ & $87.72\%$ & \textbf{84.54\%} & \textbf{79.86\%} \\
 	DeepCaps with dynamic routing & $99.19\%$ & $95.29\%$ & \textbf{91.52\%} & $87.60\%$ & $84.14\%$ & $78.66\%$ \\
		DeepCaps with self routing & $99.25\%$ & $95.60\%$ & $90.5\%$ & \textbf{88.15\%} & $83.17\%$ & $77.37\%$ \\
 \hline\hline
 \end{tabular}
 }
 \end{center}
 \label{table_affine_results_routing}
\end{table*}

We also evaluate the robustness of DeepCaps when the PGD adversarial training is applied, compared to the normally trained DeepCaps. We chose an input perturbation $\varepsilon$ equal to $0.03$, with step size $0.005$ in each iteration of the algorithm. From Figure~\ref{PGD_ADVERSARIAL_TRAINING}, we can derive that the adversarial training increases the robustness of the DeepCaps against the PGD attack, both for the CIFAR10 and GTSRB datasets, because its classification accuracy is higher than the baseline DeepCaps.

The adversarial training with PGD defense helps the networks also against the CW attack. For both the datasets, from Table~\ref{tab:Carlinin_Wagner_adversarial_training}, comparing both the mean distortion and the fooling rate, the defended DeepCaps appears more robust. Hence, the adversarial training improves the model interpretability and reduces the learning of brittle features, also when the attack algorithm used for the defense is different from the one used for the actual attack.

\section{Analyzing the Contribution of Dynamic Routing to the Robustness of the DeepCaps}
\label{sec:robustness_routing}

As a further analysis, we investigate the contribution of the dynamic routing towards the CapsNets generalization and, as a consequence, towards their robustness. 
We train two versions of the DeepCaps architecture. (i) The original dynamic routing with three iterations has been replaced by a simple connection (i.e., one iteration of dynamic routing) in both the 3D convolutional and the DigitCaps layers. (ii) The dynamic routing has been replaced by the self-routing algorithm in the last fully connected layer.
Then, we run the experiments on such networks and compare them with the original DeepCaps.

\begin{figure*}[ht!]
 \centering
 \includegraphics[width=.85\textwidth]{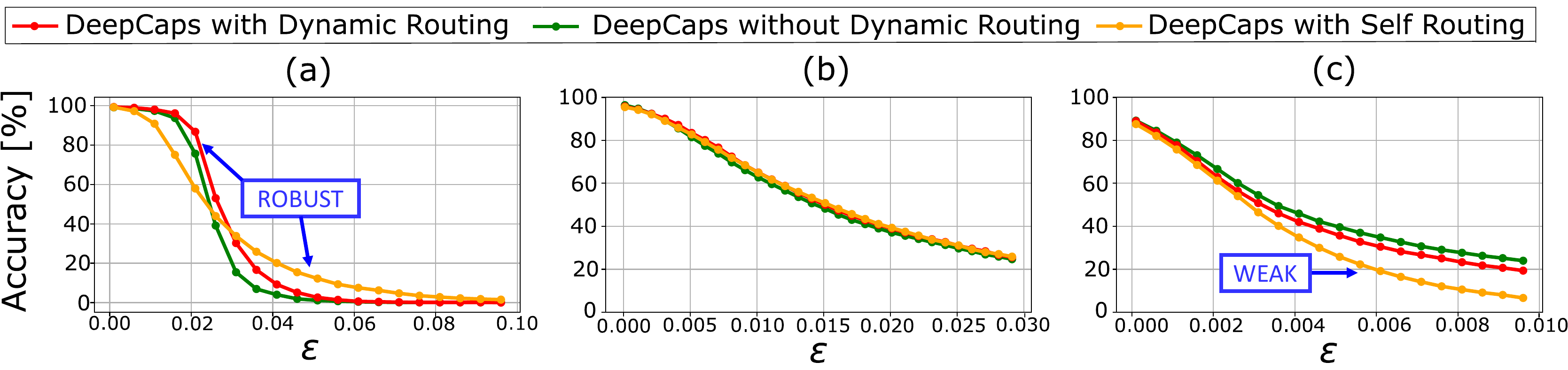}
 \caption{PGD results: comparison of the DeepCaps response for (a) MNIST and (b) GTSRB and (c) CIFAR10 datasets.}
 \label{PGD_ROUTING}
\end{figure*}

\subsection{Evaluations under the Affine Trasformations}
\label{Affine_trasformations}

The results in Table~\ref{table_affine_results_routing} compare the accuracies achieved by the DeepCaps with and without dynamic routing, and with self-routing, for the MNIST, GTSRB, and CIFAR10 datasets. 
While the difference is minimal, the response of the DeepCaps without dynamic routing against affine transformations appears to be slightly better. For the CIFAR10 dataset, even if the accuracy with the normal dataset is higher with the dynamic routing compared to the case without it, the latter works better for the affCIFAR dataset.
The self-routing shows some limits increasing the complexity of the datasets.

We can derive that the dynamic routing does not contribute significantly to the robustness against affine transformations. Indeed, it makes the DeepCaps much computationally heavier.
The functionality of the dynamic routing is to inhibit the propagation of the activation vectors with lower contribution by lowering the values of the coupling coefficients in such connections. 
Instead, the relationship between objects is learned during training by the transformation matrix, which could wrongly recognize some relationships between the inputs and a wrong output label, which the dynamic routing amplifies, together with the correct agreements. Hence, the confidence of the incorrect label increases.

\begin{table*}[t]
\centering
\caption{Robustness results against the CW attack.}
\resizebox{.95\linewidth}{!}{\begin{tabular}{c|cc|cc|cc}\hline\hline
 & \multicolumn{2}{c|}{MNIST} & \multicolumn{2}{c|}{GTSRB} & \multicolumn{2}{c}{CIFAR10} \\ 
Network & Mean Distortion & Fooling Rate & Mean Distortion & Fooling Rate & Mean Distortion & Fooling Rate\\\hline\hline
DeepCaps with dynamic routing & $1.24$ & $86.8\%$ & $1.16$ & $98.8\%$ & $0.34$ & $100\%$ \\
DeepCaps without dynamic routing & $1.62$ & $74.0\%$ & $1.27$ & $84.11\%$ & $0.46$ & $100\%$ \\
DeepCaps with self routing & \textbf{2.28} & \textbf{48.6\%} & $1.02$ & \textbf{54.4\%} & \textbf{0.52} & \textbf{99.2\%} \\ \hline \hline
\end{tabular}
}
\label{Carlini_Wagner_Routing}
\end{table*}

\subsection{Evaluations under the Adversarial Attacks}
\label{Adversarial_attackss}

The comparison analysis for the PGD attack applied to the MNIST, GTSRB, and CIFAR10 datasets are shown in Figures~\ref{PGD_ROUTING}a,~\ref{PGD_ROUTING}b and~\ref{PGD_ROUTING}c, respectively. 

For the MNIST dataset, the DeepCaps with dynamic routing is slightly more robust than the version without it. On the contrary, for the CIFAR10, the accuracy of the DeepCaps without dynamic routing decreases faster when increasing the perturbation $\varepsilon$.
We can conclude that increasing the complexity of the dataset, from MNIST toward the CIFAR10, the dynamic routing does not improve the classification capability when the input starts to be perturbed.
 
Table~\ref{Carlini_Wagner_Routing} shows the results of the CW attack.
The self-routing seems to confer robustness with this attack, even if the architecture with dynamic routing is again outperformed by the one without it.
Since the fooling rate is lower and the mean distortion is higher without dynamic routing, we can derive that the dynamic routing does not improve the robustness against such an attack. It confirms that the dynamic routing does not contribute much to the generalization.

\section{Conclusion}

In this paper, we proposed a methodology to systematically analyze the robustness of CapsNets against affine transformations and adversarial attacks.
Comparing CapsNets and CNNs, we investigated which differences play critical roles in increasing the robustness. The ShallowCaps are more robust than comparable CNNs. However, despite the high cost of training many parameters, they do not generalize well on more complex datasets. 
The analysis results demonstrate that they are more robust against adversarial attacks but show their limits against affine transformations.
Going deeper, the DeepCaps model reduces this problem, decreasing the gap between the transformed and untransformed versions of the datasets, despite the lower number of parameters. 
Against the adversarial attacks, the DeepCaps does not reach the same robustness as the ShallowCaps for a simple task like the MNIST classification. However, for a more complex dataset like the CIFAR10, their performances overcome not only a CNN with a similar architecture but also the ResNet20. In addition, the DeepCaps offers even higher robustness when the adversarial training is employed. The same conclusion can be obtained for the affine transformations, where the DeepCaps reaches a higher accuracy than the ResNet20 with the affCIFAR dataset.
Moreover, our results show that the dynamic routing does not contribute much to improving the CapsNets' robustness.

Our thorough analysis paves the way for future CapsNet designs, allowing designers to take into account adversarial attacks when targeting safety-critical applications, as well as leading the path for new adversarial attacks against CapsNets.

\section*{Acknowledgment}

This work has been supported in part by the Doctoral College Resilient Embedded Systems, which is run jointly by the TU Wien’s Faculty of Informatics and the UAS Technikum Wien.
This work was also supported in parts by the NYUAD Center for Artificial Intelligence and Robotics (CAIR), funded by Tamkeen under the NYUAD Research Institute Award CG010, the NYUAD Center for Interacting Urban Networks (CITIES), funded by Tamkeen under the NYUAD Research Institute Award CG001, and the NYUAD Center for CyberSecurity (CCS), funded by Tamkeen under the NYUAD Research Institute Award G1104.

\begin{refsize}
\bibliographystyle{ieeetr}
\bibliography{main.bib}
\end{refsize}

\end{document}